\documentclass[11pt,a4paper]{article}
\usepackage[hyperref]{acl2021}
\usepackage{times}
\usepackage{latexsym}

\usepackage{multirow}
\usepackage{arydshln}
\usepackage{graphicx}

\usepackage{microtype}

\usepackage{xcolor}
\usepackage[normalem]{ulem}
\definecolor{forestgreen}{RGB}{34,139,34}

\def\VR#1{{\color{magenta}VR: \it #1}}
\def\VRdel#1{\bgroup\markoverwith{\textcolor{magenta}{\rule[0.5ex]{2pt}{1pt}}}\ULon{#1}}

\def\DMHdel#1{\bgroup\markoverwith{\textcolor{blue}{\rule[0.5ex]{2pt}{1pt}}}\ULon{#1}}

\def\KSdel#1{\bgroup\markoverwith{\textcolor{forestgreen}{\rule[0.5ex]{2pt}{1pt}}}\ULon{#1}}

\def\IKdel#1{\bgroup\markoverwith{\textcolor{teal}{\rule[0.5ex]{2pt}{1pt}}}\ULon{#1}}

\newcommand{\personachat}{PersonaChat}
\newcommand{\topicalchat}{TopicalChat}
\newcommand{\bertscore}{BERTScore}
\newcommand{\dataset}{OTTers}
\newcommand{\alphanlg}{$\alpha$NLG}
\newcommand{\ignore}[1]{}
\newcommand{\vgpt}{vGPT2}
\newcommand{\multigen}{MultiGen}
\newcommand{\alphanlgft}{\alphanlg{}ft}
\newcommand{\indomain}{\texttt{id}}
\newcommand{\outofdomain}{\texttt{ood}}
\newcommand{\tabitem}{~~\llap{\textbullet}~~}

\aclfinalcopy 
\def\aclpaperid{2922} 

\title{{\includegraphics[scale=0.2]{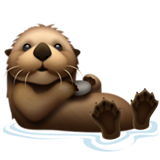} \dataset }: \\ One-turn Topic Transitions for Open-Domain Dialogue}

\author{Karin Sevegnani, David M. Howcroft, Ioannis Konstas, Verena Rieser \\
  The Interaction Lab, MACS 
    Heriot-Watt University \\
    Edinburgh, Scotland, UK \\
  \texttt{\{karin.sevegnani, i.konstas, v.t.rieser\}@hw.ac.uk} \\
  }

\date{}

\begin{document}
\maketitle
\begin{abstract}
Mixed initiative in open-domain dialogue requires a system to pro-actively introduce new topics. 
The one-turn topic transition task explores how a system connects two topics in a cooperative and coherent manner.
The goal of the task is to generate a ``bridging'' utterance connecting the new topic to the topic of the previous conversation turn.
We are especially interested in commonsense explanations of how a new topic relates to what has been mentioned before.
We first collect a new dataset of human one-turn topic transitions, which we call \dataset\footnote{https://github.com/karinseve/OTTers}.
We then explore different strategies used by humans when asked to complete such a task, and notice that the use of a bridging utterance to connect the two topics is the approach used the most.
We finally show how existing state-of-the-art text generation models can be adapted to this task and examine the performance of these baselines on different splits of the \dataset\ data.
\end{abstract}

\ignore{
tl;dr: we are not necessarily better than other datasets. We provide the way for true mixed-initiative dialogue. Our task setup is not natural per se. It was not our goal. It was designed to promote mixed-initiative and therefore teach a model how to do it proactively. So, let's avoid trying to prove that our task is natural, but instead let's try to justify that it serves a particular purpose, backed by James, 1995 (see below and lines 117-119 on the overleaf manuscript). (Potential bad analogy: GuessWhat?! dialogues, are not real dialogues, but they help a model learn object attributes.)
 
the longer version (grab a cup of tea/coffee):
After reading the paper several times, I believe we shouldn't argue that much about how bad other datasets are at doing their transitions. As Verena points out about the 'cats' example in Table 1, the first transition is not bad, the rest are weird, sure, but not for reasons we tackle, and to be honest we shouldn't even care about.
What we should point out is that if you are interested in making your conversational agent truly mixed-initiative-capable USE OTTers. Yes, there are other ways to transition topics, like acknowledge and continue, or not even mention an entity  (like the Table 1 example). But a truly engaging pseudo sentient-agent that can perform rudimentary abductive inference between two topics, needs to use a connective/transitional middle sentence with an entity related  both with Topic A/Sentence A and Topic B/Sentence B. Why? Because similar to the Cause-Effect relationship in ART (the αNLG dataset), the middle sentence (and the entity mentioned therein) of OTTers is required to provide the grounding/rationale->"bridging" between the two topics. Is this the only way to transition topics? No. Is it a better way? Yes! According to James, 1995: "each topic has a tendency to lead to the next; to provide the opening for another". 
Now, is the task artificial? Yes, partly. Why? Because we are not interested in spontaneous chit-chat dialogue: that's when you end up with Alana-style ("yeah", "sure", "I don't know") or the ConvAI2-style abrupt transitions. We are interested in KG-grounded "commonsense explanations" as Verena nicely points out in one of the comments.}

\section{Introduction}
\label{sec:intro}

\begin{figure}[t]
\small
\begin{tabular}{|rr@{~}p{3.9cm}|}
\hline
\textbf{User A} & \textbf{Source Topic}:    & I spend a lot of time \textbf{outside}.\\ \hdashline
\multirow{2}{*}{\textbf{User B}} &\textbf{Transition}:
& I like the outdoors as well, especially \textbf{\underline{gardening}}. It destresses me.\\
&\textbf{Target Topic}:    & I enjoy relaxing and getting \textbf{flowers}.\\
\hline
&\textbf{Entity Path}: & \mbox{\textbf{\texttt{outside}} - \textbf{\underline{\texttt{garden}}} -} \textbf{\texttt{flower}}\\
\hline
\hline
\textbf{User A} & \textbf{Source Topic}:    & I like \textbf{seafood} a lot. \\ \hdashline
\multirow{2}{*}{\textbf{User B}} &\textbf{Transition}:& 
Since you like seafood, is \textbf{\underline{Swedish fish}} a candy that you might enjoy?\\
&\textbf{Target Topic}:    & I have no self control when it comes to \textbf{candy}.\\
\hline
&\textbf{Entity Path}: & \mbox{\textbf{\texttt{seafood}} - \textbf{\underline{\texttt{Swedish fish}}} -} \textbf{\texttt{candy}}\\
\hline
\hline
\textbf{User A} & \textbf{Source Topic}:    & I think I am getting \textbf{engaged} soon.\\ \hdashline
\multirow{2}{*}{\textbf{User B}} &\textbf{Transition}:& 
I have two children from a previous \textbf{\underline{marriage}}\\
&\textbf{Target Topic}:    & My \textbf{children} are my life.\\
\hline
&\textbf{Entity Path}: & \textbf{\texttt{engagement}} - \textbf{\underline{\texttt{marriage}}} - \textbf{\texttt{child}}\\
\hline
\end{tabular}
\centering
\caption{
Example topic transitions from \dataset. 
User A introduces a topic with a short sentence (main concept in \textbf{bold}). Then User B responds with a (optionally multi-sentence) ``bridging'' transition before introducing the new topic (the main concepts for the transition and target topic are denoted with \textbf{\underline{underline}} and \textbf{bold}, respectively). Each example is accompanied by an entity path, comprising Knowledge Graph entities (denoted with \texttt{teletype}) instantiating the main concepts of the dialogue turn.
}
\label{img:shift_ex}
\end{figure}

For a conversation to be truly engaging, we typically assume that both participants take initiative, 
e.g.\ by introducing a new topic. We call this a {\em mixed-initiative} dialogue.
Open-domain systems trained on vast amounts of data 
\cite{jiang2020pednet, zhang2019dialogpt, gao2018neural, li2017adversarial, li2016persona, vinyals2015neural}, however, are often purely responsive, make abrupt transitions, or fail to take initiative (see examples in Table~\ref{tab:abrupt_shifts}). 
In this paper, we consider the case where the system pro-actively introduces a new topic in a conversation 
by providing a \textit{commonsense link} of how this new topic relates to what was mentioned previously (see Fig.\ref{img:shift_ex}). We call this transition strategy ``bridging''. 
Humans deploy a range of strategies in addition to bridging, including disengagement, discourse markers or silence \cite{riou2015methodology}.
We hypothesise that introducing a new topic by making a connection with the previous dialogue turn 
can be perceived as a less abrupt transition. 

More specifically, we investigate bridging transitions between two user utterances in the form of one or more sentences that contain at least one main \textit{linking} concept. These inherently can allow for better grounding to external resources such as entities in large Knowledge Graphs (KG) (e.g., Wikidata), or named entities mentioned in documents (e.g., Wikipedia, or news articles), ultimately leading to more controlled and interpretable outputs.

To this end, we crowdsource a corpus of human-written topic transitions focused on these ``bridging'' strategies, 
where humans introduce a ``missing link'' concept, given a source and target topic in the form of two short user utterances (Fig.\ \ref{img:shift_ex}). 
By grounding the topics on a KG using automatically recognised entities associated with each topic, we can then identify ``commonsense'' connections which are similar to these missing links.

By modelling such topic transitions in the form of Cause-Effect relationships in a KG, we can then perform abductive inference on commonsense knowledge for which we provide a language generation baseline.
In particular, we fine-tune a multi-hop reasoning model \cite{ji2020language}
which was trained on a similar task called Abductive NLG (\alphanlg) to generate an explanatory hypothesis given two observations. 
We find that combining a reasoning module over a KG (ConceptNet) with a language model achieves the best performance on our ``topic transition'' task for both the predicted entity path as well as the generated utterance. 
In addition, we show that existing multi-topic dialogue datasets, such as \personachat\ \cite{zhang2018personalizing} and \topicalchat~ \cite{gopalakrishnan2019topical}, cannot be easily adapted to this task,
due to the different nature of the tasks they were designed for.
Our contributions are as follows:

\begin{itemize}
 \setlength\itemsep{0.1em}
 \vspace{-1ex}
    \item We propose a new Natural Language Generation task based on one-turn topic transitions for open-domain dialogue based on a ``bridging'' strategy, which promotes grounding on KG entities.
    \item We collect a crowdsourced dataset, \dataset, and present a rigorous analysis in terms of transition strategies, linguistic properties and entity linking to a KG.
    \item We show that our KG-grounded dataset can effectively leverage the reasoning component of an existing Transformer-based model \cite{ji2020language} to generate better output compared to a vanilla GPT-2 \cite{radford2019language} decoder, both in in-domain and out-of-domain data splits.
\end{itemize}


\section{Related Work}
\label{sec:related_work}

\paragraph{Topic Transitions in the  Linguistic Literature.}
There is no common definition for the term \emph{topic} \cite{goutsos1997modeling,purver2011incremental};
however, there are a number of definitions which are helpful for our purposes.
\citet{goutsos1997modeling} divide a ``topic'' into two main components: 
1) what constitutes a topic (the ``what'') and 
2) how participants perceive and manage a topic (the ``how'').
An early work from \newcite{brown1983discourse} declares that ``topics should be described as the most frequently used, unexplained term in the analysis of discourse''.
In general, ``discourse topics'' can be explained as what a portion of the interaction is about, therefore the ``aboutness'' \citep{berthoud1995traitement, porhiel2005marqueurs}.
More specifically \newcite{chafe1994discourse} defines the notion of topic as ``the totality of information that is semiactive at one time''.


Prior work has shown that the introduction of a new topic usually co-occurs with cues such as wrapping things up about the current topic \cite{maynard1980placement},
preceding silence, or 
the use of discourse markers  \cite{riou2015methodology}. 
Also, backchannel signals, e.g., \emph{yeah}, \emph{right}, \emph{you know}, indicate that both agents are involved in the interaction and show consent for the topic development \cite{james1995topic}.
Beyond these overt cues, \citet{james1995topic} and \citet{geluykens1993topic} describe semantic topic transitions:
``each topic has a tendency to lead to the next; to provide the opening for another'' \cite{james1995topic}, and 
topics are typically ``co-constructed'', requiring each speaker to contribute to the conversation for further progression and development \cite{geluykens1993topic}.
The identification of topic transition is indeed not an easy task.
It is not only about linguistic cues such as discourse markers and prosodic cues, as sometimes a topic switch can be identified with the introduction of a new entity \cite{james1995topic}.
Additionally, in a conversation topics are created and introduced by participants themselves in real time, making topics participant- and interaction-specific \cite{mondana2001gestion, mondada2003talk}.
Moreover, ``the entities in focus at a given point in the discourse will be that partially-ordered subset of activated entities which are likely to be continued as topics of subsequent utterances'' \citep{gundel1993cognitive}.
These cooperative elements emphasise the importance of mixed-initiative topic management for open-domain dialogue systems.


\paragraph{Current Multi-topic Open-domain Systems.}

\begin{table}[h]
\centering
\footnotesize
\begin{tabular}{|r@{~}p{5.4cm}|}
\hline
PersonaChat  &\\
\hline
\textbf{User A}: & I do not like carrots. I throw them away.\\
\textbf{User B}: & Really. But, I can sing pitch perfect. \\
\textbf{User A}: & I also cook, and I ride my bike to work.\\
\textbf{User B}: & Great! I had won an award for spelling bee.\\
\hline
\hline
TopicalChat & \\
\hline
\textbf{User A}: & Yeah and saltwater fish are lucky because they can do that and drink through their mouths. \\
\textbf{User B}: & Seems like fresh water fish got the short end of the stick with that one. Have you ever been to a cat cafe? \\
\hline
\end{tabular}
\caption{Examples of abrupt topic transitions from the PersonaChat and TopicalChat datasets.}
\label{tab:abrupt_shifts}
\end{table}

Previous work in open-domain dialogue systems has largely avoided explicitly modelling topic transitions and instead focused on grounding system behaviour in a ``persona'' (a set of statements about hobbies, demographics, or preferences) \cite{zhang2018personalizing, li2016persona} or by conditioning conversations on knowledge sources such as newspaper articles, fun facts or Wikipedia articles \cite{gopalakrishnan2019topical, dinan2018wizard} to generate engaging responses while avoiding generic replies, improving coherence, and raising new and interesting topics. 
These approaches often lead to poor topic transitions, as illustrated in Table \ref{tab:abrupt_shifts}. 
The PersonaChat example shows neither initiative nor common sense while transitioning to a new topic; it only displays passive acknowledgement from User B.
Whereas the TopicalChat example presents a very abrupt topic shift by User B.
Our dataset is the first corpus focused specifically on one-turn topic transitions;
however, there are several human-to-human dialogue corpora wherein participants discuss assigned topics.
Two prominent such corpora are \topicalchat\ \citep{gopalakrishnan2019topical} and \personachat\ \citep{zhang2018personalizing}.

In \topicalchat\ both participants used source documents from Wikipedia to discuss a shared topic.
The dialogues in this corpus tend to flow less naturally than those in \personachat\,
with participants 
generally focusing on expressing the main facts, often by copy and pasting from their source documents
rather than having a natural conversation.
Therefore we focus on \personachat\ as a point of comparison.
%

\personachat\ dialogues consist of chit-chat conversations based on a set of ``persona traits'' assigned to each participant.
Because participants seek to express their persona to each other, 
the conversations require mentioning various topics (i.e.\ their persona traits) in a natural way.
Indeed, \citet[Sec.\ 3.3]{zhang2018personalizing} adjusted their design to encourage users to engage with each other's topics and not simply state their own topics as quickly as possible to end the dialogue.
\personachat\ does not contain annotations for the topic of each turn and participants had the freedom to mention their topics (i.e.\ persona traits) in any order.

We use \personachat\ in two different ways: 
1) using their persona traits as starting and goal topics for our own data collection, and 
2) as a point of comparison for our dataset.

\paragraph{Commonsense-Aware Neural Text Generation.}
Large Language Models still suffer in cases where reasoning over underlying commonsense knowledge is  required  during  generation, including dialogue generation \cite{ijcai2018-643},  story ending generation \cite{Guan_Wang_Huang_2019}, and  
topic-to-essay generation \cite{yang-etal-2019-enhancing-topic}. Recently, \citet{Guan_Wang_Huang_2019,bhagavatula2019abductive} attempted to integrate external commonsense knowledge into generative pretrained language models, which we will also attempt in Section \ref{sec:experiments} using the Abductive NLG (\alphanlg) dataset \cite{bhagavatula2019abductive}.
Our setup is similar in spirit to \alphanlg, which is a conditional generation task for explanations given observations in natural language.  In particular, the 
model has to generate an explanatory hypothesis given two observations: the cause (e.g. {\em The Smith family went on a cruise for their summer vacation}) and the consequence (e.g. {\em From then on, the Smiths went to the beach each summer instead}). 
Here, a possible explanation might be: {\em The Smith family got seasick on the cruise}.
The \alphanlg\ dataset contains $20$k pairs observations and $200$k explicative hypotheses, which we will later use for fine-tuning our models (see Section \ref{sec:experiments}).


\section{One-turn Topic Transitions}
\label{sec:ottrs}


\subsection{Task Design and Data Collection}
\label{sec:ottrs:task-design}

\paragraph{Task Description.} We assume there are topics $t_a$ and $t_b$ for utterances $u_a$ and $u_b$ (with $u_{\cdot} = t_{\cdot}$ for this paper).
The goal of the task is to generate a \emph{one-turn transition} utterance $u_t$ to serve as a smooth link between $t_a$ and $t_b$ so that its concatenation with utterance $u_b$ is a sensible response to $u_a$.
A \emph{bridging} transition occurs when one or more of the entities $e_t \in \mathbf{e_t}$ mentioned in $u_t$ lies on a path in the knowledge graph between entities $e_a \in \mathbf{e_a}$ and $e_b \in \mathbf{e_b}$ mentioned in $u_a$ and $u_b$, respectively.

\paragraph{Knowledge Graph Construction.}
We use \personachat\ persona traits as the starting point for our data collection. In order to model commonsense connections,
we built a knowledge graph (KG) using the entities found in each persona trait through the Yahoo Entity Linker \cite{Blanco:WSDM2015, Pappu:WSDM2017}.
Each entity is linked to its correspondent Wikidata identifier, while a SPARQL query retrieved the entity's super-classes and sub-classes, 
which were added to the KG. 
Furthermore, the KG has been augmented by retrieving the commonsense connections for each entity from ConceptNet \citep{speer2017conceptnet} and by parsing Wikipedia abstracts mentions.

To select which traits to use for the data collection, we first selected all pairs of entities connected with $k$-hops ($1<k<20$) in the KG.
Then, we recovered the entities mentions in the persona traits and saved every pair (nearly $30$k) as potential pairs for our data collection.

\begin{figure}[h]
\includegraphics[width=0.45\textwidth]{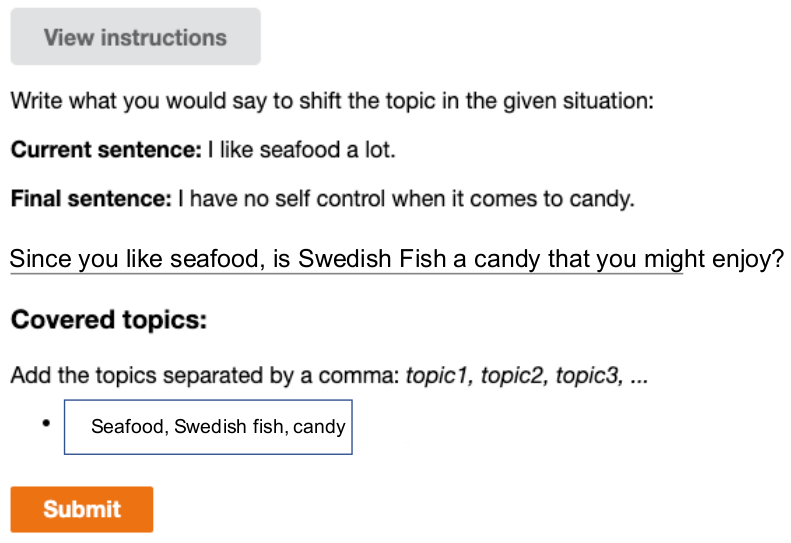}
\centering
\caption{Screenshot of part of the interface users are presented with when accepting the study on AMT.}
\label{img:amt}
\end{figure}

\paragraph{Data Collection.}
We crowdsourced the data collection for \dataset\ on Amazon Mechanical Turk (AMT). 
Each user was provided with two topics \textbf{A, B} from the \personachat\ persona traits, 
along with instructions explaining the task.
The instructions ask the user to imagine they are having a conversation 
where the first topic \textbf{A} from the pair represents the last turn of the other person, 
and the second topic \textbf{B} contains the final topic the user wants to talk about.
The user then has to write a short utterance to transition to the new topic \textbf{B} in the least abrupt way possible.
Additionally, in order to encourage crowd-workers to ground their utterances in actual topics,
we asked them to report the ``topics'' mentioned in their sentence (see Figure \ref{img:amt}).

For each topic pair in the study we collected three different transition utterances to provide more insight into the different strategies users adopt when transitioning to a new topic.


\subsection{Corpus Properties}
\label{ssec:otters:properties}

\paragraph{Basic Statistics.}
Table \ref{tab:corpus-properties} provides summary statistics describing \dataset. 
Our corpus consists of $4{,}316$ utterances for $1{,}421$ unique topic pairs, with an average utterance length of $1.3$ sentences and $16.4$ words.
The KG path statistics for \dataset\ are based on all of the paths found by the Yahoo Entity Linker between the $1421$ unique topic pairs in the corpus, a total of just over $12k$ paths.


\begin{table}[]
\small
    \centering
        \begin{tabular}{|l|r|}
        \hline
        Property  & \textbf{\dataset} \\
        \hline
        \textbf{TTR}       & $5085$ / $70935$\\
        \hline
        \textbf{ent.\ TTR}  &  $3984$ / $17054$\\
        \hline
        \textbf{turn length}  &  $2$-$68$ \\
        ---: mean    & $16.4$\\
        ---: mode    &  $9$\\
        \hline
        \textbf{sent.s/turn}  &  $1$-$7$\\
        ---: mean    &  $1.3$\\ 
        ---: mode    &  $1$\\
        \hline
        \textbf{KG path}      &  $1$-$14$\\
        ---: mean    &   $6.1$\\
        ---: mode    &   $5$\\
        \hline
    \end{tabular}    
    \caption{Properties of \dataset. 
    \emph{Type-token ratio} (TTR) for the different splits of the dataset. \emph{Entity TTR} refers to the number of (unique) entities appearing in that portion of the dataset. \emph{Turn length} is the length of the transition utterance in tokens, including punctuation. \emph{Number of sentences per turn} measured by splitting on sentence-final punctuation.}
    \label{tab:corpus-properties}
\end{table}

\paragraph{KG coverage.}
We calculated the distance between each pair of topics in the knowledge graph described in Sec.\ \ref{sec:ottrs:task-design} to facilitate analyses of the role of topic distance in transition strategy and transition quality.
To extract entities from the utterances in our corpus, 
we extended the tagger built-in to the Yahoo Entity Linker with the spaCy Named Entity Recognizer to include all nouns and adjectives as potential entities.\footnote{This modified version allowed us to identify a wider range of topic-related entities.} 



Using these extracted entities we analyse the overlap between entities mentioned in the given topics \textbf{A, B} and those mentioned in the crowdsourced transition utterances.
The Jaccard distance between these two sets is $1$ for nearly a quarter of the topic-pairs and utterances in our dataset, with a mean of $0.842$, meaning that the overlap between entities mentioned in the utterances and entities mentioned in the topics is fairly low. 
This indicates that users transition from Topic A to Topic B mentioning new unseen entities, following a ``path'' that can be grounded on a knowledge graph.

In contrast, the overlap between the entities in the KG path between the topics and the entities mentioned in the transition utterances is higher:
both the mean and the mode Jaccard distances drop to below $0.8$, 
suggesting that crowdworkers make similar connections to the ones we can find in our knowledge graph a substantial portion of the times.
This suggests that our KG-grounded approach can find plausible entities to be mentioned to bridge between topics, similar to the commonsense connections made by humans shifting between topics.


\begin{table}[]
\small
\centering
\begin{tabular}{|l|c|c|}
\hline
              &  \multicolumn{2}{c|}{Jaccard Dist.} \\
     overlap  &  Mean                & Mode (freq)           \\
     \hline
utt.-topics   & $0.842$             & $1.0$ ($1274$)    \\
utt.-KG path  & $0.751$             & $0.667$ ($451$)     \\     
\hline
\end{tabular}
\label{tab:ents}
\caption{Overlap in entities between transition utterances and (1) the topic sentences (i.e.\ persona traits) and (2) the path between those topics in the KG.}
\end{table}

\begin{table}[]
\small
\centering
\begin{tabular}{|l|l|l|}
\hline
            & Spearman $\rho$ & Pearson $r$\\
\hline
Cambridge   & $0.039*$    &     $0.046*$    \\ 
PDTB3       & $0.003$      &    $0.001$      \\ 
turn length & $0.139$      &   $-0.001$    \\ 
\hline
\end{tabular}
\caption{Correlations between the KG distance between topics \textbf{A, B} and the number of discourse markers used, as defined by \emph{Cambridge Dictionary} and \emph{Penn Discourse Treebank}, as well as correlation with turn length.
* indicates statistical significance $p<0.01$ with Bonferroni correction for multiple comparisons.}
\label{tab:correlations}
\end{table}

\subsection{Transition Strategies in \dataset}
\label{sec:validating:transition-strategies}

To examine the strategies humans applied while completing the \dataset\ task, we adapted the categories of \citet{riou2015methodology} for a manual analysis of our data.
\citet{riou2015methodology} distinguishes between \emph{disjunctive} and \emph{stepwise} transitions between topics.
Disjunctive transitions make no attempt to relate the new topic to the previous topic,
switching abruptly to the new topic without acknowledging the previous topic,
whereas stepwise transitions are akin to the previously described transition strategies.

We distinguish between \emph{bridging} and \emph{acknowledge \& continue} strategies:
in the former, the speaker aims to produce an utterance which connects the previous and new topics directly;
in the latter, the speaker acknowledges the previous topic before introducing their own topic, without explicitly relating the two to one another.
In addition to these categories, we also annotated utterances as \emph{off-task} (e.g. replying to or continuing the first topic without any attention paid to the second topic) or \emph{off-topic} when the utterance had nothing to do with either of the two topics 
(e.g.\ random greetings or generic questions).

Two of the authors annotated $10$ utterances from $10$ different users, resulting in $200$ total annotations.
The initial inter-annotator agreement was $71\%$, classified as substantial (Krippendorff's $\alpha = 0.34$), after which the annotators collaborated to reach a consensus annotation for each of the examples that presented a disagreement.
Table \ref{tab:example-transitions} contains a prototypical example for each of the annotated classes.

More than $80\%$ of the data contains some form of transition to the second topic, 
with $79\%$ containing a bridging utterance, $5\%$ applying an acknowledge and continue strategy, and only $2\%$ using the disjoint transition strategy.
$12\%$ of the data is connected to one or more of the topics in some way but does not serve as a transition, 
and $2\%$ of the data is completely off-topic.
This analysis suggests that our corpus indeed represents the kind of knowledge-based transitions we are interested in.  


\paragraph{KG distance and discourse markers.}
We hypothesize that speakers are less likely to use explicit topic management strategies (e.g.\ topic wrap-ups, discourse markers) when topics are more closely related to each other, e.g.\ as measured by graph distance in a large knowledge graph.
This would be in line with findings about the use of explicit discourse markers versus leaving discourse relations implicit.
\citet{torabi-asr-demberg-2012-implicitness,torabi-asr-demberg-2013-information} 
found that explicit markers are more likely to be omitted when the discourse relation is highly predictable based on the content of the arguments. 

Based on \citet{riou2015methodology} we examined the frequency of discourse markers in utterances to test our hypothesis,
examining both general conversational discourse markers and those associated with specific discourse relations. 
For conversational discourse markers we use the Cambridge Dictionary, which provides a list of spoken and written markers, including 
``well'', ``you know'', etc.,
while for markers signalling particular discourse relations we use the list from the Penn Discourse Treebank \cite[PDTB]{webber.etal2019, prasad-etal-2008-penn}; these include markers like ``because'' indicating a causal relationship or ``in addition'' for an additive relationship.
We find a small but significant correlation ($\approx 0.04$) between conversational discourse markers and no significant correlations between the use of PDTB3 discourse markers or the turn length and KG distance. 
This suggests that users are somewhat more likely to use conversational discourse markers as the distance between topics in the knowledge graph increases, in line with our hypothesis.


\begin{table}[]
    \centering
    \small
    \begin{tabular}{|rp{6.5cm}|}
         \hline
         \multicolumn{2}{|c|}{\textbf{Acknowledge and continue}}\\
         \hline
         A: & i like to eat the same thing as ninja turtles.\\
         T:  & I love pizza.  I eat it while I skateboard.\\
         B: & i enjoy riding around on a plank with wheels.\\
         \hline
         \multicolumn{2}{|c|}{\textbf{Bridging: Missing Link}}\\
         \hline
         A: & i prefer things to be authentic.\\
         T:  & I think children are the truest form of authenticity because they say things unfiltered.\\
         B: & i am not a fan of children.\\
         \hline
         \multicolumn{2}{|c|}{\textbf{Disjunctive}}\\
         \hline
         A: & i like american made cars.\\
         T:  & I like liver cooked in butter -- just throwing that in!\\
         B: & i avoid eating broccoli.\\
         \hline
         \multicolumn{2}{|c|}{\textbf{Off-Task}}\\
         \hline
         A: & i prefer things to be authentic.\\
         T:  & my bro just made some authentic thai chicken.\\
         B: & i am not a fan of children.\\
         \hline
         \multicolumn{2}{|c|}{\textbf{Off-Topic}}\\
         \hline
         A: & i learnt to drive.\\
         T:  & I had a rough night sleeping in my new bed last night.\\
         B: & i like making a salmon entree.\\
         \hline
    \end{tabular}
    \caption{Prototypical examples of each annotation category for transition strategy or lack of transition in \dataset. (A) is the preceding topic or utterance; (T) is the collected utterance; and (B) is the goal topic and potential next utterance for Speaker B.}
    \label{tab:example-transitions}
\end{table}

\subsection{Validating the Corpus}
\label{sec:validating-the-corpus}

We evaluate whether the transition strategies in \dataset\ are less abrupt than those found in \personachat
by constructing a comparable subset of \personachat and  performing a human evaluation.

\ignore{
\subsection{Transition Strategies in \dataset}
\label{sec:validating:transition-strategies}
\VR{Move this under Section \ref{ssec:otters:properties}. It's not validation, but more a data description.}
To examine the strategies humans applied to the \dataset\ task, we adapted the categories of \cite{riou2015methodology} for a manual analysis of our data.
\citet{riou2015methodology} distinguishes between \emph{disjunctive} and \emph{stepwise} transitions between topics.
Disjunctive transitions make no attempt to relate the new topic to the previous topic,
switching directly to the new topic without acknowledging the previous topic.

We view stepwise transitions as more cooperative, and we distinguish between the \emph{bridging} and \emph{acknowledge \& continue} strategies.
In a bridging transition, the speaker aims to produce an utterance which connects the previous and new topics directly.
In contrast, they might \emph{acknowledge} the previous speaker's topic before introducing their own topic, without clearly relating the two to one another.

In addition to these categories, we also annotated utterances as potentially \emph{off-task} (e.g. replying to or continuing the first topic without any attention paid to the second topic) or \emph{off-topic} when the utterance had nothing to do with either of the two topics provided.
(e.g.\ random greetings or generic questions).

The first two authors annotated 10 utterances from 10 different users, resulting in 200 annotations total.
Initial interannotator agreement was 71\% (Krippendorff's $\alpha = 0.34$), after which the two annotators collaborated to reach a consensus annotation for each of these 100 examples.
The consensus distribution of labels is shown in Figure \ref{fig:annotation-distribution}.


More than 80\% of the data contains some form of transition to the second topic, 
with 79\% containing a bridging utterance, 5\% applying an acknowledge and continue strategy, and only 2\% using the disjoint transition strategy.
12\% of the data is connected to one or more of the topics in some way but does not serve as a transition, 
and 2\% of the data is completely off-topic.
Of the 79 bridging transitions, only 3 used the so-called `summarising' strategy, 
suggesting that our corpus indeed represents the kind of transitions we want our models to learn.
Table \ref{tab:example-transitions} contains examples from the annotated portion of the dataset.

\begin{table}[]
    \centering
    \small
    \begin{tabular}{rp{6.5cm}}
         \hline
         \multicolumn{2}{c}{\textbf{Acknowledge and continue}}\\
         \hline
         A: & i like to eat the same thing as ninja turtles.\\
         T:  & I love pizza.  I eat it while I skateboard.\\
         B: & i enjoy riding around on a plank with wheels.\\
         \hline
         \multicolumn{2}{c}{\textbf{Bridging: Missing Link}}\\
         \hline
         A: & i prefer things to be authentic.\\
         T:  & I think children are the truest form of authenticity because they say things unfiltered.\\
         B: & i am not a fan of children.\\
         \hline
         \multicolumn{2}{c}{\textbf{Disjunctive}}\\
         \hline
         A: & i like american made cars.\\
         T:  & I like liver cooked in butter -- just throwing that in!\\
         B: & i avoid eating broccoli.\\
         \hline
         \multicolumn{2}{c}{\textbf{Off-Task}}\\
         \hline
         A: & i prefer things to be authentic.\\
         T:  & my bro just made some authentic thai chicken.\\
         B: & i am not a fan of children.\\
         \hline
         \multicolumn{2}{c}{\textbf{Off-Topic}}\\
         \hline
         A: & i learnt to drive.\\
         T:  & I had a rough night sleeping in my new bed last night.\\
         B: & i like making a salmon entree.\\
    \end{tabular}
    \caption{Prototypical examples of each annotation category for transition strategy or lack of transition in \dataset. (A) is the preceding topic or utterance; (T) is the collected utterance; and (B) is the goal topic and potential next utterance for Speaker B.}
    \label{tab:example-transitions}
\end{table}
}

\paragraph{Comparable Corpus Construction.}
\label{sec:comparable_corpus_construction}
We first extract a subset of  \personachat\ where two consecutive turns contain different topics. 
In other words: turns where one speaker changed the topic from what the previous speaker has just said.
Since \personachat\ turns do not incorporate topic annotations, 
we use a heuristic based on \bertscore\ to assign a topic to each turn.
Given topics $\mathbf{t}$ and turns $\mathbf{u}$ for a dialogue in \personachat, 
we calculate the \bertscore\ similarity between each $u \in \mathbf{u}$ and each $t \in \mathbf{t}$.
For each turn $u$ we then assign $t = \textrm{argmax}_t(\textrm{\bertscore}(u, t))$, if and only if 

\begin{equation}
    \frac{\textrm{\bertscore}(u, t)}{\textrm{\bertscore}(u, t')} > d
\end{equation}

where $t'$ is the topic achieving the second highest \bertscore\ relative to $u$, and $d$ is a threshold 
to ensure that we only assign a topic to a turn if it is a substantially better fit than 
the other topics.\footnote{We set $d = 1.27$ for our dataset construction, since this is the 50th percentile of \bertscore\ values observed in our data.}
While this means that 
not every turn is assigned a topic, this is necessary to ensure that we do not assign topics to, e.g., greetings like `hi, how are you?'.

This way of assigning topics yields a subset consisting of $22{,}010$ utterances which have a different topic from the preceding utterance.
Most of these topic-pairs ($20{,}491$) are only expressed through one utterance in the dataset, while $1{,}188$ are expressed by two utterances, $248$ by three, and $83$ by more than $3$ utterances.
Moreover, there are $445$ topic-pairs which also occur in our corpus. 

\paragraph{Crowdsourced Validation.}
\label{sec:validation:human-eval}

\begin{figure}
    \centering
    \includegraphics[trim=0.08cm 0 0 0,clip,width=0.5\textwidth]{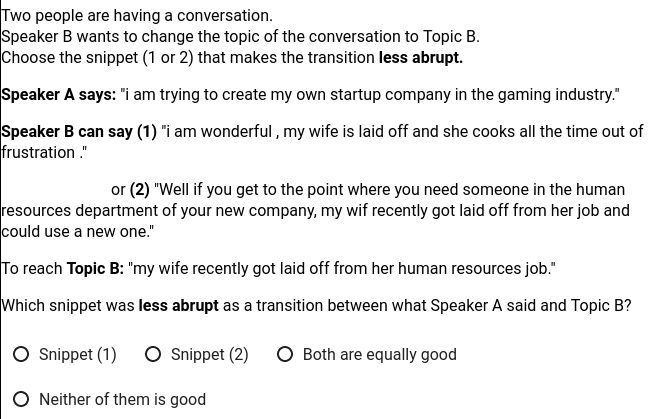}
    \caption{Interface for crowdsourced validation.}
    \label{fig:human-eval}
\end{figure}

Using the comparable sub-corpus of \personachat,  
we asked crowdworkers to vote which of two potential transition utterances was ``less abrupt'' (Fig.\ \ref{fig:human-eval}) for $49$ topic-pairs occurring in both datasets.
We collected $3$ votes for each utterance and only counted instances where 2/3 workers agreed on the same choice.

The results confirm that \dataset\ has less abrupt transitions:
the utterances in \dataset\ were judged as less abrupt in 44/49 cases, with the comparable \personachat\ utterance judged less abrupt in one case, and both utterances rated ``bad'' in another. 
Only 3 cases did not present a majority class.

\section{Experiments}
\label{sec:experiments}
Having confirmed the quality of our corpus, we now adapt two existing text generation models 
as baselines for this task.
We also explore different train-dev-test splits and conduct an error analysis. 

\subsection{Baselines} 
\label{sec:experiments:baselines}

The first baseline we consider is a vanilla GPT-2 language model \citep{radford2019language} fine-tuned on \dataset\  (\vgpt). 
Next, we  test the recent \multigen\ \cite{ji2020language} on this task, which extends GPT-2 with multi-hop reasoning on commonsense knowledge graphs. 
In particular, this model combines the vocabulary distribution generated by GPT-2 with a concept distribution in order to produce knowledge grounded responses.
The concept distribution is given by reasoning performed on the commonsense knowledge graph ConceptnetIO, using the context modeled through GPT-2.

\subsection{Train-Dev-Test Splits}
\label{sec:experiments:splits}

The first split is an out-of-domain split (\outofdomain), which ensures that none of the topics in the test-set are present in any of the topic-pairs in the train-set.
For the second split, this restriction is relaxed to create an in-domain split (\indomain), allowing one of the topics in each pair in the test-set to appear in the train-set, although with a different second topic.

The \outofdomain\ split resembles a zero-shot scenario, where the model has to generate a shift between two topics it has never been fine-tuned on.
Hence, we expect results to be lower than the ones from \indomain.
The number of unique and total topic pairs for each split is illustrated in Table \ref{tab:data_split}.

\begin{table}[]
\centering
\begin{tabular}{|l|c|c|c|}
\hline
              & Train     & Dev      & Test      \\
\hline
\indomain     & $693$/$1{,}929$  & $404$/$1{,}160$ & $303$/$1{,}158$  \\
\outofdomain & $677$/$2{,}034$  & $372$/$1{,}152$ & $372$/$1{,}130$ \\
\hline
\end{tabular}
\caption{Num.\ unique/total topic-pairs in each split.}
\label{tab:data_split}
\end{table}

\subsection{Evaluation}
\label{sec:experiments:evaluation}

We evaluate two aspects of the transition task:
1) whether the model can find a sensible path through intermediate topics and
2) whether the model can generate a natural utterance which mentions such intermediate topics.


To evaluate the former, we assess the entities mentioned in the transition utterance to determine how well they bridge the gap between Topic A and Topic B.
We use \texttt{hits@k} ratio as an automatic approximation, which
measures the number of relevant entities correctly predicted by the model, out of the \texttt{k} most important entities identified in the target references.
This metric shows how well the models 
ground the concepts introduced in the two dialogue turns and how the reasoning compares to the human standard presented in \dataset.

For (2) 
we adopt the same automated metrics used for evaluating \multigen\ 
 on the \alphanlg\ dataset for comparability:
ROUGE-L \cite{lin2004rouge}, METEOR \cite{banerjee2005meteor}, and CIDEr \cite{vedantam2015cider}.
However, we report the full BLEU score \cite{papineni2002bleu}\footnote{we use SacreBLEU \cite{post-2018-call}.} that accounts for the overlap across 1-4 ngrams instead of just 4-grams (BLEU-4). 
As word-overlap based metrics have been widely criticised due to their lack of correlation with human judgements \citep{novikova2017we,reiter2018}, 
we also provide an example-based error analysis in Section \ref{sec:experiments:results}.


\subsection{Results} 
\label{sec:experiments:results}
\begin{table*}[t]
\small
\centering
\begin{tabular}{|l|l|r|r|r|r|r|r|}
\hline
        & \textbf{split}      & \textbf{BLEU} & \textbf{METEOR} & \textbf{ROUGE-L} & \textbf{CIDEr} & \textbf{hits@1} & \textbf{hits@3} \\
\hline
vGPT2 & \texttt{ood} & $1.26$ & $8.37$  & $12.4$ & $4.65$ & $22.94$  & $12.54$   \\
vGPT2 & \texttt{id} &  $1.58$   &   $10.26$     &    $14.67$  & $3.75$  &   $\mathbf{57.14}$     &   $\mathbf{30.79}$     \\
\hline
\alphanlg & \texttt{ood}& $1.52$ & $16.35$  & $23.26$ & $12.12$ & $21.85$  &  $12.03$  \\
\alphanlg & \texttt{id}  &  $1.6$    &    $18.9$    &    $25.52$  & $11.33$  & $22.89$  &  $12.72$      \\
\hline
\alphanlg{}ft & \texttt{ood}      & $4.0$  & $22.8$  & $29.27$   & $\mathbf{21.7}$ & $35.33$  & $18.67$ \\
\alphanlg{}ft & \texttt{id}    & $\mathbf{4.27}$     &   $\mathbf{25.03}$     &   $\mathbf{31.54}$      & $18.48$   &   $38.11$   &   $21.51$    \\
\hline
\end{tabular}
\caption{System performance on in-domain and out-of-domain data splits.}
\label{tab:baselines_res}
\end{table*}

\begin{table}[h]
\centering
\footnotesize
\begin{tabular}{|p{0.7cm}@{~}|rp{4.6cm}|}
\hline
& Topic A: & I like babies.\\
Ex. 1&\dataset & I make babies laugh.\\
&Topic B: & I enjoy making other people laugh.\\
\hline
&vGPT2:   & I like other people making babies.\\
&\alphanlg: & I decided to give birth to a baby.\\
&\alphanlg{}ft: & I like to make babies laugh when I'm not working.\\
\hline
\hline
&Topic A: & I love those sleek Italian cars.\\
Ex. 2&\dataset & I wish I was as sleek as those cars.\\
&Topic B: & I am in good shape.\\
\hline
&vGPT2:   & I love it. I love it.\\
&\alphanlg: & I bought a new car.\\
&\alphanlg{}ft: & I like the look of Italian cars.\\
\hline
\hline
&Topic A: & I have prior agriculture experience.\\
Ex. 3&\dataset & My parents were farmers.\\
&Topic B: & I am a mom.\\
\hline
&vGPT2:   & a lot of cooking. I am a lot of cooking.\\
&\alphanlg: & My mom taught me how to grow vegetables.\\
&\alphanlg{}ft: & I have been a farmer for many years.\\
\hline
\hline
&Topic A: & I make blogs.\\
Ex. 4&\dataset & Do you know of any blogs that are on the subject of large families with babies?\\
&Topic B: & I have a large family with babies.\\
\hline
&vGPT2:   & family.\\
&\alphanlg: & I have a lot of babies.\\
&\alphanlg{}ft: & I have a large family.\\
\hline
\end{tabular}
\caption{Representative outputs from each model.}
\label{tab:output_exs}
\end{table}

For each aforementioned split we evaluated three different models to compare performance: 
the pretrained \vgpt\ fine-tuned on each split for \dataset, 
the \multigen\ model fine-tuned only on \alphanlg, 
and the same model additionally fine-tuned on \dataset\ (called \alphanlgft).

\paragraph{Overview of Results.}
Table \ref{tab:baselines_res} shows the results of these experiments.
\vgpt\ performs poorly on the one-turn transition task, regardless of the train-dev-test split,
which we attribute to the small size of \dataset:
with only a few thousand utterances, \vgpt\ is unable to learn the task.
We notice, however, that the system tends to repeat the main entity in Topic A, therefore scoring surprisingly well on the \texttt{hits@k} metric, despite the fact that the utterances themselves are of low quality (see Table \ref{tab:output_exs}).


The reasoning component added by \multigen\ leads to substantial improvements in most of the evaluation metrics but not \texttt{hits@k} (\alphanlg\ in the table).
Therefore, the improvements in text quality metrics appear to be due primarily to the similarity between the structure of the abductive NLG task and the increased amount of data for fine-tuning ($\approx688$k tokens) compared to fine-tuning \vgpt\ on our $\approx71$k tokens alone.

Further fine-tuning \multigen\ on \dataset\ leads to substantial improvements on all metrics for both in-domain \& out-of-domain splits.
The performance improvement is considerable especially because of the relatively small size of the training set ($693$ unique topic pairs on in-domain, see Table \ref{tab:data_split}), further justifying the compatibility between the original task \multigen\ was trained on and \dataset.
Nonetheless, the BLEU scores from Table \ref{tab:baselines_res} indicate 
there is still space for improvement.
We hypothesise METEOR are higher than BLEU scores, because they also consider paraphrases.

These results confirm that our newly introduced one-turn topic transition task needs a reliable language model combined with an advanced reasoning component.

\paragraph{Detailed Discussion and Model Limitations.}
We further analyse the results 
to understand model limitations.
First, we observe that 
Multigen's \texttt{hits@k} ratio is quite low, especially when compared to 
\vgpt. 
This is surprising considering \vgpt's generated sentences are mostly very short and repetitive, and the predicted concepts mostly match the ones contained in the `Topic A' sentence. 
One possible explanation is that
Multigen's reasoning module uses a gate loss, which determines whether to select a concept from the provided knowledge graph or a word from the GPT2 dictionary.
We observed that the majority of the times the model will use a word from the GPT2 dictionary rather than selecting a concept from the knowledge graph.

Moreover, we observe that only $65\%$ of the concepts found in the target sentences are actually nodes in Multigen's subgraphs. 
One possible explanation is that Multigen's reasoning model has a limited input capacity of 
 up to 100 nodes that are at most 2 hops away
in order to prune the very large knowledge graph from ConceptNet.
The English vocabulary from ConceptNet contains approximately $1{,}500{,}000$ nodes, which makes the process of determining the concept distributions very computationally expensive and time inefficient.
Therefore, the pruning strategy adapted by \citet{ji2020language} overcomes these problems but cannot be applied to the OTTers task, as the selection of the concepts is just as important as the output sentence being fluent.
Contrary to our expectations, expanding the size of the knowledge graphs from $100$ nodes to $200$ and $300$ did not improve the \texttt{hits@k} ratio.
Most likely because the concepts added to the graphs are either not relevant or misleading for the model.
This suggests that 
improving concept selection is a promising future direction to 
improve the performance of the reasoning module, leading to overall better topic transitions.

\ignore{ 
The nice easy story we want to tell is:
    vanilla-GPT2 fine-tuned on our dataset (1st row of Table 6), performs poorly, as in it mostly copies parts of the first sentence. It is only able to recover some entities though, hence the abnormally high HITS@k. Example outputs can backup this claim.
	vanilla αNLG tested only (i.e., not trained) on our dataset (2nd row of Table 6), performs better than vanilla-GPT2, since it is fine-tuned on a similar task first and also grounded on ConceptNet. However, there is still a mismatch of language between our task and their task
	αNLG fine-tuned on our dataset (3rd row of Table 6), performs eay better than the rest, even with just 693 unique pairs of traits (a total of 1929 examples, after including multiple reference examples) of training data. This makes sense given that we are fine-tuning on our task and the boost is remarkable even with a model that is not specific to the task or encoding the entity traits explicitly. Of course, there is room for improvement given the relatively low BLEU. METEOR is much higher in comparison, possibly due to taking into account paraphrases. Karin let's quickly have a look at BLEU-4 alone, just in case there is more of a story to tell about longer sequences; ROUGE-L (longest common subsequences) is also quite high.
	The 4th row of results can be ignored.
	o-o-d vs i-d (out-of-domain vs in-domain): o-o-d resembles a zero-shot scenario where the model has to generate a shift about traits it has not been fine-tuned before on, hence scores are lower than the -id version (test set may contain at most one trait it has seen before during training). Probably reporting this number on the vanilla αNLG model (2nd row) isn't as interesting: since no fine-tuning has taken place, everything is new to this model, right?
} 

\paragraph{Error Analysis.}
In addition, we preform an example-based error analysis to further understand the strengths and weaknesses of the individual models.
Table \ref{tab:output_exs} shows representative system outputs for each of the models on the in-domain data split. 
First, we observe that 
\vgpt\ often generates very simple sentences (e.g., `\textit{family.}', in Ex. 4), repeated non-content bearing tokens (e.g., \textit{`I love it.'}, in Ex. 2), or  incoherent and often not specific enough output to form a successful bridging transition (e.g., \textit{`a lot of cooking.'}, in Ex. 3, is not a well-formed sentence, and only loosely connected to Topic A about \textit{`agricultural experience`}), contributing to low BLEU scores.
However, this also reinforces the idea that the \texttt{hits@k} scores are artificially inflated simply due to \vgpt\ choosing to include one of the entities from the first topic.

The outputs from \multigen\ tested on \dataset\ show a better performance than \vgpt, given that the topic selection for the model is grounded on ConceptNet. 
However, since the Abductive NLG task is different than the `Topic Transition' task addressed in \dataset, there is a discrepancy in the use of the language.
The model often outputs coherent sentences that use generic commonsense facts which may not be related to Topic B (e.g., `\textit{I decided to give birth to a baby'}, in Ex. 1).

The texts generated from \multigen\ fine-tuned on \dataset\, on the other hand, 
introduce interesting connections between Topic A and Topic B (e.g., `\textit{I like to make babies laugh when I'm not working.'}, in Ex. 1) and leverage commonsense (e.g., `\textit{I like the look of Italian cars'}, in Ex. 2, where \textit{`the look'} creates a connection with \textit{`being in good shape'} from Topic B).

\section{Discussion \& Conclusion} 
\label{sec:conclusions}

\paragraph{Ethical Considerations.}
We recognise that any mixed-initiative dialogue system carries risks related to dual-use:
in addition to helpful systems which serve to help users explore a new topic or discover more about the world,
a system which can effectively change the topic of conversation 
could also be used to manipulate user behaviour.
For example, bridging strategies for topic transitions could be used by virtual assistants to encourage users to make a purchase or to express their opinion or preference regarding sensitive subjects.

\paragraph{Conclusion.}
We have defined a new NLG task exploring one-turn topic transitions for mixed-initiative in open-domain systems.
Our \dataset\ corpus provides training data for modelling topic transitions based on `missing link' topics which connect the previous conversation subject to a new topic.
Baseline models based on state-of-the-art approaches to text generation illustrate possible approaches to the task and show that there is room for improvement.
In particular, we show that commonsense knowledge grounding is necessary for this task, outperforming fine-tuned large language models. In future work, we will explore model architectures specifically designed for topic transitions, as well as fine-tuning strategies to deal with small datasets.
We also plan to evaluate the impact of bridging transitions on user (dis)engagement in an open-domain dialogue system.

\section*{Acknowledgements}
This research received funding from the EPSRC project MaDrIgAL (EP/N017536/1), as well as Google Research Grant to support NLU and dialog research at Heriot-Watt University.

\bibliography{acl2021}
\bibliographystyle{acl_natbib}
\newpage
\newpage
\appendix

\section{OTTers Data Collection}
In order to avoid collecting noisy or out-of-task data, we established some worker requirements for turkers participating in our data collection.
Workers needed to:
\begin{itemize}
    \item be Masters (label assigned by Mechanical Turk to workers who achieve excellence across a variety of tasks),
    \item have a number of HITs approved greater than $500$,
    \item have a HIT approval rate ($\%$) greater than $80$,
    \item being located in an English speaking country, namely Australia, Canada, New Zealand, United Kingdom, and United States.
\end{itemize}

A Worker received a reward of $\$0.3$ for each completed assignment.
The reward was calculated based on an estimate of the time it would take a Worker to read the instructions and complete the task.
The time for completing the task has been estimated at $1.5$ minutes, and the reward was calculated accordingly to a $\$12$ hourly payment.
Each task had been assigned to 3 unique workers.

\begin{figure*}[]
    \centering
    \includegraphics[width=\textwidth]{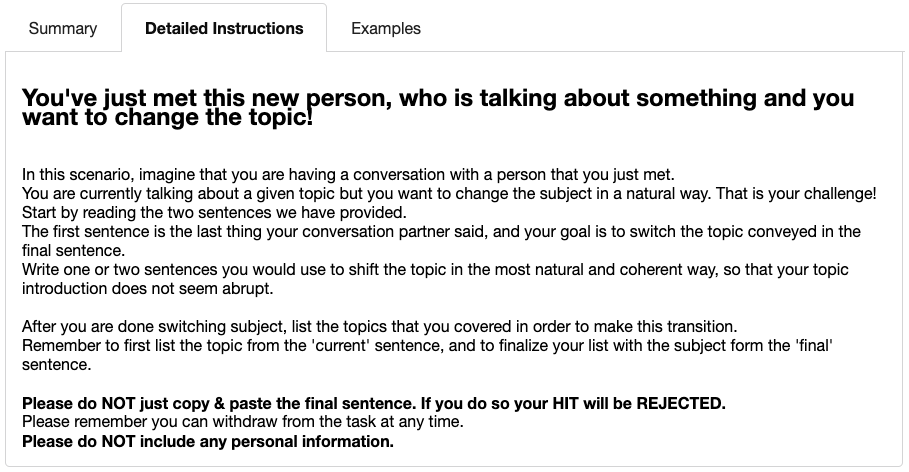}
    \caption{Instructions shown to Workers for the OTTers data collection.}
    \label{fig:amt_instr}
\end{figure*}

Figure \ref{fig:amt_instr} shows the instructions that Workers were presented after opening the OTTers data collection task.
The instructions explain that the context is a conversation with a newly-met person.
After writing the sentence for transitioning the current topic to the `final' one, workers are asked to list the topics they covered for the transition.
Additionally, we provided an example:

\textbf{Current sentence}: `I have a love of reptiles.' 

\textbf{Final sentence}: `I want to travel to NYC.'

\textbf{Topic shifting sentence}: `I know there is a cool snake species in the New York zoo. This is why I want to travel to NYC.'

\textbf{Covered topics}:
\begin{itemize}
    \item Reptiles
    \item Zoo
    \item NYC
\end{itemize}

\begin{table*}[]
\centering
\begin{tabular}{|l|l|}
\hline
\textbf{Current sentence} &  I like the pool.\\
\textbf{Topic transition} &  I reward myself after going to the pool by eating a hearty meal.\\
\textbf{Final sentence}   &  I like vegetables.\\
\hline
\hline
\textbf{Covered topics} & \tabitem Pool \\
    & \tabitem Hearty meal \\
    & \tabitem Vegetables \\
\hline
\end{tabular}
\end{table*}

\begin{table*}[]
\centering
\begin{tabular}{|l|l|}
\hline
\textbf{Current sentence} &  I am a parent.\\
\textbf{Topic transition} &  My kids each have a pet they take care of.\\
\textbf{Final sentence}   &  I like animals.\\
\hline
\hline
\textbf{Covered topics} & \tabitem Parent \\
    & \tabitem Kids \\
    & \tabitem Pet \\
    & \tabitem Animals \\
\hline
\end{tabular}
\end{table*}

\begin{table*}[]
\centering
\begin{tabular}{|l|l|}
\hline
\textbf{Current sentence} &  I have gone across the ocean.\\
\textbf{Topic transition} &  My mom was in a band when she lived in France.\\
\textbf{Final sentence}   &  My mom is famous.\\
\hline
\hline
\textbf{Covered topics} & \tabitem Ocean \\
    & \tabitem Mom \\
    & \tabitem Band \\
    & \tabitem France \\
    & \tabitem Famous \\
\hline
\end{tabular}
\end{table*}

\begin{table*}[]
\centering
\begin{tabular}{|l|l|}
\hline
\textbf{Current sentence} &  I love cuddling with my babies.\\
\textbf{Topic transition} &  I wish I could still do that with my children, but after my back injury I have had to \\
 & listen to my physician and really change my ways.\\
\textbf{Final sentence}   &  The Dr said no sitting up for me.\\
\hline
\hline
\textbf{Covered topics} & \tabitem Children \\
    & \tabitem Medical care \\
    & \tabitem Disability \\
\hline
\end{tabular}
\end{table*}

\begin{table*}[]
\centering
\begin{tabular}{|l|l|}
\hline
\textbf{Current sentence} &  I like going to concerts.\\
\textbf{Topic transition} &  I like going to concerts which means I normally have to take a break during a workweek.\\
\textbf{Final sentence}   &  I do not go a full week of employment without a break.\\
\hline
\hline
\textbf{Covered topics} & \tabitem Concerts \\
    & \tabitem Break \\
    & \tabitem Work \\
\hline
\end{tabular}
\end{table*}

\begin{table*}[]
\centering
\begin{tabular}{|l|l|}
\hline
\textbf{Current sentence} &  I play basketball and football.\\
\textbf{Topic transition} &  I bet my fiancee while at the park playing basketball.\\
\textbf{Final sentence}   &  My significant other and I will be having a wedding.\\
\hline
\hline
\textbf{Covered topics} & \tabitem Basketball \\
    & \tabitem Football \\
    & \tabitem Fiancee \\
    & \tabitem Park \\
    & \tabitem Wedding \\
\hline
\end{tabular}
\end{table*}

\end{document}